\documentclass{article}

    \PassOptionsToPackage{numbers, compress}{natbib}
 \usepackage[preprint]{neurips_2026}

\usepackage[utf8]{inputenc} % allow utf-8 input
\usepackage[T1]{fontenc}    % use 8-bit T1 fonts
\usepackage{hyperref}       % hyperlinks
\usepackage{url}            % simple URL typesetting
\usepackage{booktabs}       % professional-quality tables
\usepackage{amsfonts}       % blackboard math symbols
\usepackage{nicefrac}       % compact symbols for 1/2, etc.
\usepackage{microtype}      % microtypography
\usepackage{xcolor}         % colors

\usepackage{graphicx}
\graphicspath{{lorec/}} % 告诉 LaTeX 去 lorec 文件夹里找图
\usepackage{caption}      % 基础标题控制
\usepackage{subcaption}   % 子图支持（包含 \captionsetup[subfigure] 的定义）
\usepackage{wrapfig}
\usepackage{amsmath}
\usepackage{amssymb} % 建议同时加载，用于支持 \mathbf 等符号
\usepackage{multirow}
\usepackage{booktabs} % 建议同时加载，用于 \toprule, \midrule, \bottomrule
\usepackage[table]{xcolor} % 关键：[table] 选项必须有
\usepackage{algorithm}
\usepackage{algorithmic}
\usepackage{makecell}

\usepackage{hyperref}       % 必须先加载 hyperref

\usepackage[capitalize]{cleveref} % 必须后加载 cleveref
\title{LoReC: Rethinking Large Language Models for Graph Data Analysis}

\author{%
  \textbf{Hongyu Zhan\textsuperscript{1}, Qixin Wang\textsuperscript{2}, Yusen Tan\textsuperscript{1}, Haitao Yu\textsuperscript{1}, Jingbo Zhou\textsuperscript{3}} \\
  \textbf{Shuai Chen\textsuperscript{4}, Jia Li\textsuperscript{1}, 
  Xiao Tan\textsuperscript{4}\thanks{Corresponding authors.}, 
  Jun Xia\textsuperscript{1,5,$\dagger$}} \\  
  \textsuperscript{1}The Hong Kong University of Science and Technology (Guangzhou) \\
  \textsuperscript{2}Tongji University, \textsuperscript{3}Westlake University, \textsuperscript{4}Ant Group \\
  \textsuperscript{5}The Hong Kong University of Science and Technology \\
  \texttt{\{hzhan701, ytan277, hyu382\}@connect.hkust-gz.edu.cn} \\
  \texttt{2450850@tongji.edu.cn, zhoujingbo@westlake.edu.cn} \\
  \texttt{\{shuai.cs, alex.tx\}@ant-intl.com} \\
  \texttt{\{jialee, junxia\}@hkust-gz.edu.cn} \\
}

\begin{document}

\maketitle

\begin{abstract}
  The advent of Large Language Models (LLMs) has fundamentally reshaped the way we interact with graphs, giving rise to a new paradigm called GraphLLM. As revealed in recent studies, graph learning can benefit from LLMs. However, we observe limited benefits when we directly utilize LLMs to make predictions for graph-related tasks within GraphLLM paradigm, which even yields suboptimal results compared to conventional GNN-based approaches. Through in-depth analysis, we find this failure can be attributed to LLMs' limited capability for processing graph data and their tendency to overlook graph information. To address this issue, we propose \textbf{LoReC} (\textbf{Lo}ok, \textbf{Re}member, and \textbf{C}ontrast), a novel plug-and-play method for GraphLLM paradigm, which enhances LLM's understanding of graph data through three stages: (1) \textbf{Look}: redistributing attention to graph; (2) \textbf{Remember}: re-injecting graph information into the Feed-Forward Network (FFN); (3) \textbf{Contrast}: rectifying the vanilla logits produced in the decoding process. Extensive experiments demonstrate that LoReC brings notable improvements over current GraphLLM methods and outperforms GNN-based approaches across diverse datasets. The implementation is available at \url{https://github.com/Git-King-Zhan/LoReC}.
  % https://anonymous.4open.science/r/LoReC-63F4
  % https://github.com/Git-King-Zhan/LoReC
\end{abstract}

\section{Introduction}

Recently, the integration of Graph Neural Networks (GNNs) with Large Language Models (LLMs) has catalyzed a new paradigm termed GraphLLM, empowering LLMs to understand complex graph data and conduct downstream graph tasks~\cite{huang2024can, chen2024llaga, he2025unigraph}. In many cases, existing works have practically demonstrated that LLMs can effectively facilitate graph learning. For example, GraphGPT~\cite{tang2024graphgpt} aligns graph structural knowledge with LLMs for graph-related tasks, and Dr.E~\citep{liu2025multi} utilizes LLMs to comprehend graph data. Nevertheless, we observe minor improvements or even significant performance degradation when directly utilizing LLMs for predictions in graph-related tasks within GraphLLM paradigm. Previous work~\cite{li2024glbench} also observed unsatisfactory performance when utilizing LLMs to make predictions on graphs, but did not delve into this issue.

\begin{wrapfigure}{r}{0.5\textwidth} 
  \centering
  \vspace{-10pt} % 调整图片与上方正文的间距
  \captionsetup[subfigure]{
    labelfont=normalfont,
    size=footnotesize, % 环绕布局空间有限，建议缩小字号
    skip=2pt,
    margin=0pt
  }
  % 第一行
  \begin{subfigure}[b]{0.23\textwidth} 
    \centering
    \includegraphics[width=\linewidth]{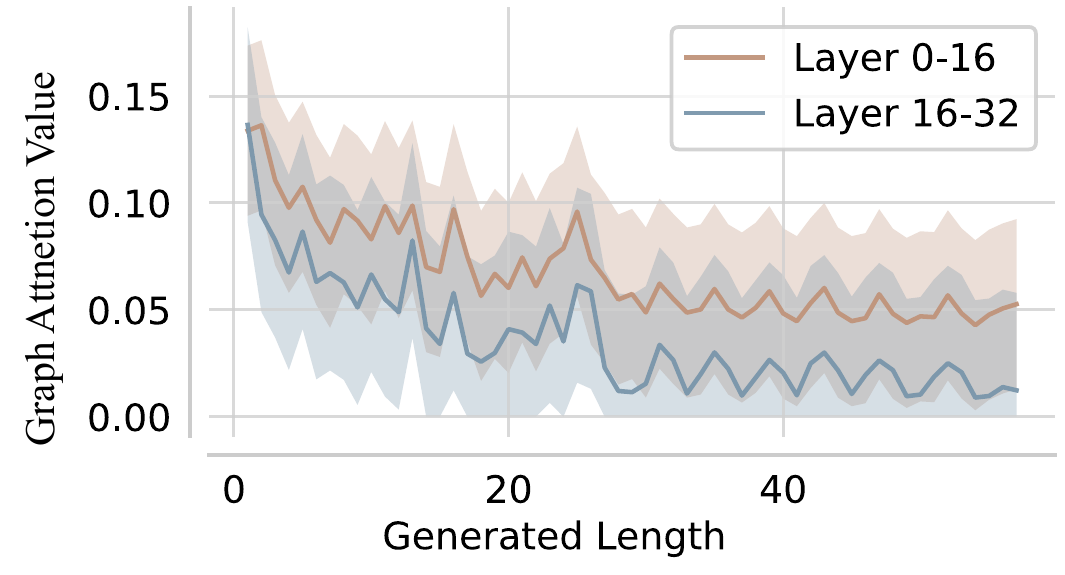}
    \caption{Graph Tokens} 
    \label{fig:a}
  \end{subfigure}
  \hfill
  \begin{subfigure}[b]{0.23\textwidth}
    \centering
    \includegraphics[width=\linewidth]{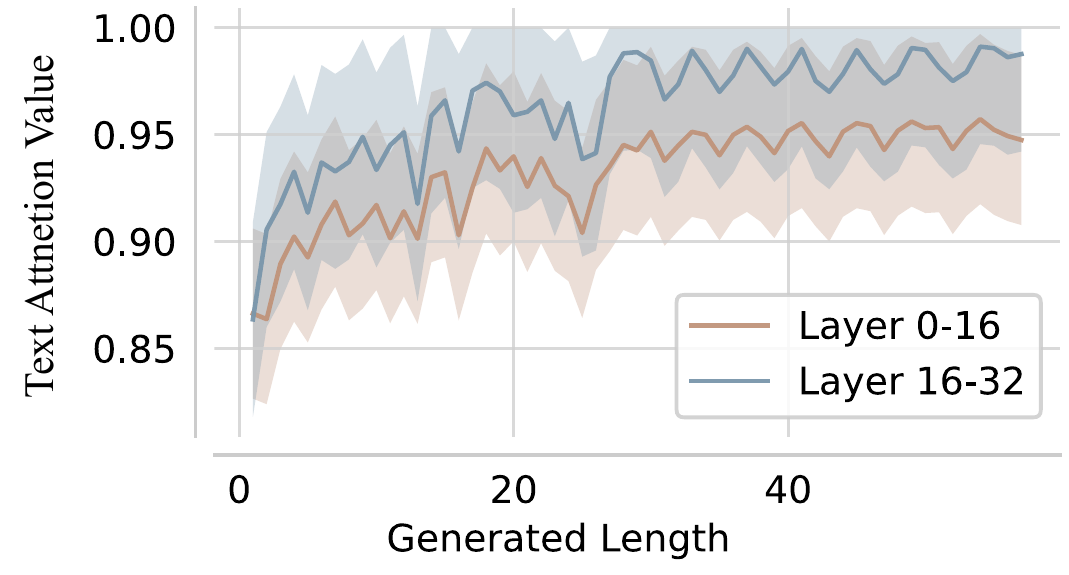}
    \caption{Text Tokens} 
    \label{fig:b}
  \end{subfigure}

  \vspace{4pt} 
  % 第二行
  \begin{subfigure}[b]{0.23\textwidth}
    \centering
    \includegraphics[width=\linewidth]{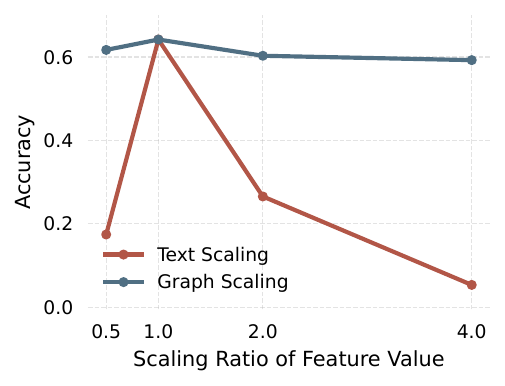}
    \caption{Scaling} 
    \label{fig:c}
  \end{subfigure}
  \hfill
  \begin{subfigure}[b]{0.23\textwidth}
    \centering
    \includegraphics[width=\linewidth]{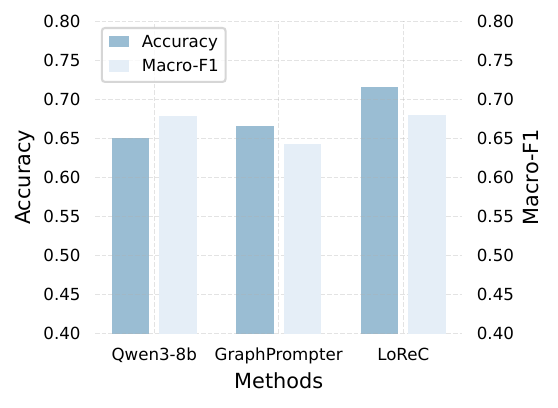}
    \caption{Performance Comparison} 
    \label{fig:d}
  \end{subfigure}
  \caption{(a-b) Attention distribution change between graph and text tokens across decoding layers as generation length increases. (c) Model performance under different scaling ratios applied to graph and text features. (d) Performance comparison of three approaches on the Citeseer dataset: text-only LLM baseline (Qwen3-8b), GraphPrompter, and LoReC-enhanced GraphPrompter.}
  \label{modality_imbalance}
  \vspace{-10pt} % 调整图片与下方正文的间距
\end{wrapfigure}

To explain these phenomena, we conduct several experiments as shown in \cref{modality_imbalance}. As depicted in \cref{fig:a} and \cref{fig:b}, attention to text tokens increases while attention to graph tokens decreases as more tokens are generated, impeding LLMs' understanding of graph data. This issue worsens in deeper layers, where over $90\%$ of attention is allocated to text tokens. To investigate the respective contributions of graph-structured data and textual context to predictions, we proportionally scale text and graph feature values. As shown in \cref{fig:c}, performance degradation from scaling text features is more pronounced than from scaling graph features, indicating that GraphLLM predictions are predominantly driven by textual features. Moreover, as demonstrated in \cref{fig:d}, text-only LLM achieves performance comparable to GraphLLM model despite the absence of graph inputs, revealing that predictions are primarily driven by textual priors rather than graph inputs. Additional evidence across more datasets is provided in the Appendix \ref{pubmed}.

As a consequence, performance degradation in GraphLLM models stems from two interrelated factors: \textit{diluted attention allocated to graph tokens} and \textit{overconfidence in textual priors}. Specifically, certain text tokens with minimal predictive utility receive disproportionately high attention weights during decoding, which intensifies as decoding proceeds stepwise. Furthermore, the higher information density inherent in graph-structured data impedes GraphLLM models from effectively comprehending graph inputs, causing these models excessively rely on textual priors. This naturally motivates the following question: \textit{Can we enhance LLMs' understanding of graph data as well as mitigate their over-reliance on textual priors?}

To answer this question, we propose \textbf{LoReC} (\textbf{Lo}ok, \textbf{Re}member, and \textbf{C}ontrast), a novel training-free decoding framework for the GraphLLM models, inspired by the natural human cognition process: \textit{when attempting to understand complex information, people instinctively redistribute attention to overlooked evidence, calibrate their memory, and rectify conflicting information before committing to an answer}. Following this principle, we employs a three-stage pipeline: (\textbf{\textit{i}}) attention redistribution to ensure the model ``\textbf{looks}" at graph tokens, (\textbf{\textit{ii}}) memory calibration via graph information reinjection to ensure it ``\textbf{remembers}" graph evidence, and (\textbf{\textit{iii}}) dual-contrastive decoding to ensure it ``\textbf{contrasts}" away false priors. The main contributions can be summarized as follows:
\begin{itemize}
    \item We conduct comprehensive and in-depth investigation demonstrating why GraphLLM models exhibit suboptimal performance.
    \item We propose LoReC, a three-stage training-free decoding framework that enhances GraphLLM models' understanding of graph data.
    \item Extensive experiments across diverse datasets demonstrate that LoReC brings consistent and notable improvements over state-of-the-art GraphLLM methods and outperforms conventional GNN-based approaches.
\end{itemize}

\section{Related Work}
\subsection{Large Language Models}
The emergence of Large Language Models (LLMs) has revolutionized deep learning across multiple domains, demonstrating remarkable capabilities in understanding and generation. Leading models, notably ChatGPT~\cite{achiam2023gpt} and Gemini~\cite{comanici2025gemini}, have pushed the frontier of artificial intelligence with advanced capabilities in multimodality, long-context processing and agentic reasoning. Concurrently, the open-source ecosystem has thrived, with powerful foundation models such as Llama~\cite{grattafiori2024llama}, Qwen~\cite{yang2025qwen3} and Gemma~\cite{team2025gemma}. A significant trend is the enhancement of complex reasoning. For instance, DeepSeek-R1~\cite{guo2025deepseek} demonstrates that reasoning capabilities can be effectively incentivized through reinforcement learning, complementing prompting strategies like Chain of Thought~\cite{wei2022chain} and Tree of Thoughts~\cite{yao2023tree}. Underpinning these successes are foundational techniques such as In-Context Learning~\cite{dong2024survey} for few-shot generalization and Reinforcement Learning from Human Feedback(RLHF)~\cite{ouyang2022training} for instruction alignment. Despite these remarkable strides in text processing and logical reasoning, LLMs exhibit inherent limitations when interpreting complex structural data. A recent benchmark~\cite{dai2025how} investigates how LLMs understand graph patterns and reveals that LLMs struggle to understand graph topologies and structural patterns. This critical gap has motivated the development of GraphLLM paradigm, which aims to equip LLMs with graph comprehension capabilities.

\subsection{GraphLLM}
Recently, the convergence of Graph Neural Networks (GNNs) and Large Language Models (LLMs) has established the nascent field of GraphLLM. This paradigm fundamentally transforms graph interaction by enabling the alignment between graph structures and natural languages. Initial efforts addressed the gap by projecting graph components into the LLM's token space via instruction tuning, as exemplified by GraphGPT~\cite{tang2024graphgpt} and LLaGA~\cite{chen2024llaga}, soft prompting strategies like GNP ~\cite{tian2024graph} and GraphPrompter~\cite{liu2024can}, or efficient adapter modules as seen in GraphAdapter~\cite{huang2024can}. To achieve finer granularity and broader generalization, subsequent research introduced token-level quantization techniques, such as Dr.E~\cite{liu2025multi}, and developed unified graph foundation models. Additionally, GOFA~\cite{kong2024gofa} interleaves GNN layers with LLMs, while Unigraph~\cite{he2025unigraph} leverages text-attributed graphs for cross-domain transfer. Furthermore, to address the context window limitations inherent in large-scale graphs, GraphChain~\cite{wei2025graphchain} innovates by treating the LLM as an intelligent agent that performs analysis via logical tool chaining. However, benchmarks including GLBench~\cite{li2024glbench} and GraCoRe~\cite{yuan2025gracore} indicate that despite promising performance in semantic understanding and zero-shot scenarios, the misalignment between graph topology and text semantics often impedes accuracy in downstream tasks.

\section{Background}
\label{motivation}
\subsection{Preliminaries}
\textbf{Graph-Structure Data.} A graph is defined as $\mathcal{G} = (\mathcal{V}, \mathcal{E}, \mathbf{A}, \mathbf{X})$, where $\mathcal{V}$ is the set of $N$ nodes and $\mathcal{E}$ represents the edges. The topological structure is encoded by an adjacency matrix $\mathbf{A} \in \{0, 1\}^{N \times N}$, where $\mathbf{A}_{ij} = 1$ indicates a connection between nodes $v_i$ and $v_j$. Each node $v_i$ is associated with a feature vector $\mathbf{x}_i \in \mathbb{R}^F$, forming the feature matrix $\mathbf{X} \in \mathbb{R}^{N \times F}$. 
% In the context of Text-Attributed Graphs (TAGs), $\mathbf{X}$ is often derived from the semantic encoding of textual attributes associated with each node.

\textbf{Graph Neural Networks (GNNs).} GNNs learn node representations by recursively aggregating information from local neighborhoods. This process, referred to as the \textit{message-passing paradigm}, distinguishes graph representation learning from architectures designed for regular modalities such as images or text. The layer-wise update for a node $v$'s embedding $\mathbf{h}_v^{(l)}$ is formally defined as:
\begin{equation}
    \mathbf{h}_v^{(l+1)} = \psi^{(l)} \left( \mathbf{h}_v^{(l)}, \phi^{(l)} \left( \left\{ \mathbf{h}_u^{(l)} : u \in \mathcal{N}(v) \right\} \right) \right),
\end{equation}
where $\mathcal{N}(v)$ denotes the neighbors of $v$. Here, $\phi(\cdot)$ serves as the aggregation function (pooling messages from neighbors), and $\psi(\cdot)$ acts as the update function to fuse the aggregated context with the node's current state. After $L$ layers, the final representations $\mathbf{h}^{(L)}$ are utilized for downstream tasks such as classification or link prediction.

\textbf{Autoregressive Language Models.} LLMs are typically instantiated as autoregressive transformer decoders trained via Causal Language Modeling (CLM). Formally, given a sequence of tokens $\mathbf{s} = (s_1, s_2, \dots, s_R)$, the model estimates the joint probability $p(\mathbf{s})$ by decomposing it into a sequence of conditional probabilities:
\begin{equation}
    p(\mathbf{s}) = \prod_{t=1}^{R} p(s_t \mid s_{<t}),
\end{equation}
where $s_{<t}$ denotes the preceding context. The core mechanism driving this estimation is \textit{masked self-attention}, which allows the model to attend to historical tokens while preventing information leakage from future tokens. For a specific attention head $h$, the output $\mathbf{O}_h$ is computed as:
\begin{equation}
    \mathbf{O}_h = \text{softmax}\left(\frac{\mathbf{Q}_h \mathbf{K}_h^\top}{\sqrt{d_k}}\right) \mathbf{V}_h,
\end{equation}
where $\mathbf{Q}_h, \mathbf{K}_h, \mathbf{V}_h$ serve as the query, key, and value matrices projected from the input embeddings, and $d_k$ is the scaling factor. The final probability distribution over the vocabulary is obtained by applying a softmax function to the projected hidden states of the last layer.

\subsection{Problem Formulation}
Given a GraphLLM model $\mathcal{M_\theta}$ parameterized by $\theta$, with a general architecture consisting of a text embedding layer, a graph encoder, a graph-text interface module, a text decoder with $L$ transformer layers, and an affine layer $\varsigma(\cdot)$ that predicts the next-token distribution. For a graph-grounded task with textual query $q$ and input graph $\mathcal{G}$, GraphLLM models first utilize the graph encoder $f_{\mathcal{G}}(\cdot)$ to transform the raw graph structure into a sequence of latent embeddings $\mathbf{Z}_{\mathcal{G}} = f_{\mathcal{G}}(\mathcal{G}) \in \mathbb{R}^{N \times d_g}$, where $N$ is the number of nodes and $d_g$ is the embedding dimension. The graph embeddings $\mathbf{Z}_{\mathcal{G}}$ are then projected into the LLM's word embedding space $\mathbb{R}^{d_{llm}}$ as graph context tokens $\mathbf{C}_{\mathcal{G}}$, which are concatenated with the textual instruction tokens $\mathbf{C}_{\mathcal{T}}$ and autoregressively decoded into the textual response $\mathbf{Y} = \{y_1, y_2, \dots, y_{max}\}$:
% \vskip -0.15in
\begin{equation}
    p(\mathbf{Y} \mid \mathcal{G}, \mathbf{H}_{\mathcal{T}}) = \prod_{t=1}^{max} p(y_t \mid \mathbf{H}_{\mathcal{G}}, \mathbf{H}_{\mathcal{T}}, y_{<t}; \theta),
\end{equation}
    % \vskip -0.1in
where $y_{<t}$ denotes previously generated tokens and $\theta$ represents the model parameters. This paradigm enables GraphLLM models to perform graph-related tasks.

\begin{figure*}[t] % <--- 改为 figure*，通常建议位置参数设为 [t] (页顶)
  \centering
    \centering
    \includegraphics[width=\linewidth]{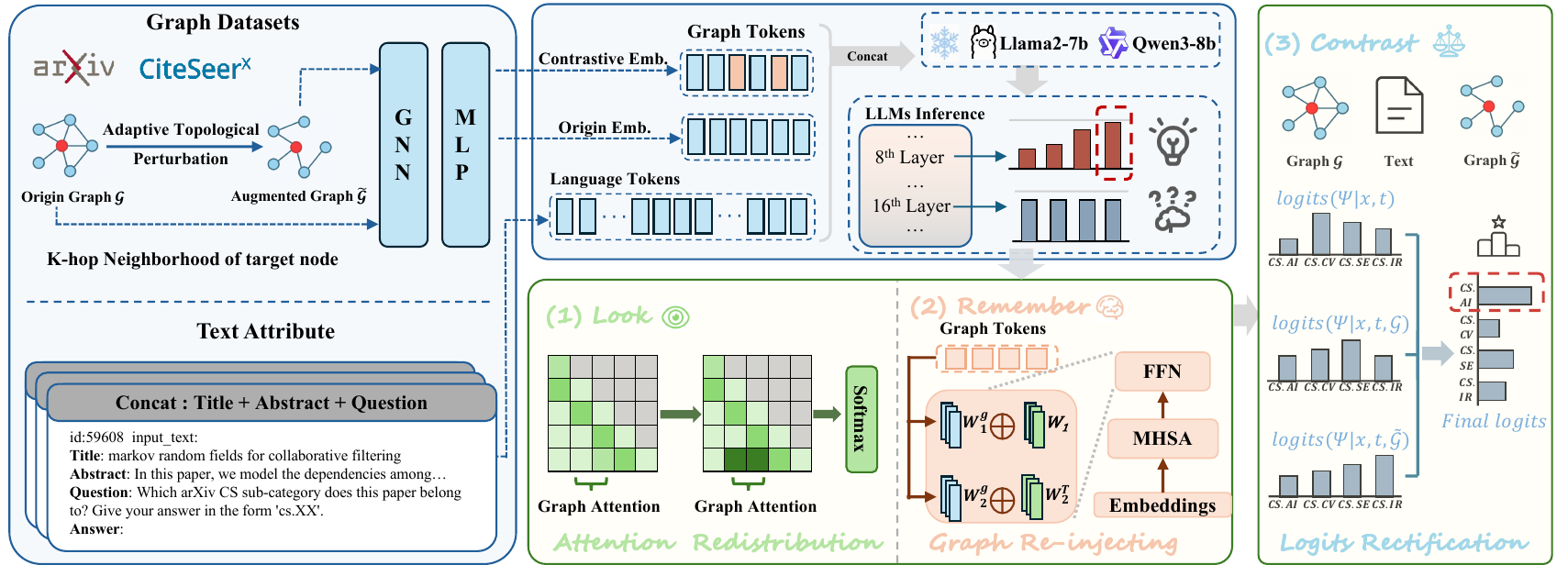}
  \caption{The overall framework of our proposed LoReC.}
  \label{framework}
\end{figure*}

\section{Methodology}
Building on the background outlined in Sec. \ref{motivation}, we propose \textbf{LoReC}, a framework containing three major components: (\textbf{\textit{i}}) attention redistribution to ensure the model ``\textbf{looks}" at graph tokens, (\textbf{\textit{ii}}) Graph Re-injection via graph information reinjection to ensure it ``\textbf{remembers}" graph evidence, and (\textbf{\textit{iii}}) dual-contrastive decoding to ensure it ``\textbf{contrasts}" away false priors. We present the framework overview in \cref{framework} and all details are listed below.

\subsection{\underline{\textbf{\textit{Look}}}: Attention Redistribution}
\label{sec:Attention_Redistribution}

As demonstrated in \cref{modality_imbalance}, GraphLLM models allocate progressively less attention to graph tokens in deeper layers as the generation process proceeds. This phenomenon causes models to overlook the graph tokens $\mathbf{C}_{\mathcal{G}}$, preventing them from effectively ``looking'' at the graph data during inference. To counteract this graph attention dilution problem, we propose a mechanism dubbed \textbf{Attention Redistribution} that dynamically amplifies attention to graph tokens based on predictive uncertainty. The overall procedure is outlined in the Appendix \ref{pseudo}.

\textbf{Uncertainty-Aware Trigger.} We employ the Shannon entropy of the next-token distribution as a proxy for model uncertainty, which is computed by a vocabulary head $\varsigma(\cdot)$ on each layer during decoding~\cite{chuang2024dola}. Formally, let $P_{\theta}(i \mid \mathbf{x}_{<t}, \mathcal{G})$ denote the probability distribution over the top $N$ candidates at time step $t$, conditioned on the preceding context $\mathbf{x}_{<t}$ and the graph input $\mathcal{G}$. The predictive uncertainty $\mathcal{H}_t^{(l)}$ at layer $l$ is calculated as:
\begin{equation}
    \label{eq:uncertainty_entropy}
    \mathcal{H}_t^{(l)} = - \frac{1}{\log N} \sum_{i=1}^{N} P_{\theta}(i \mid \mathbf{x}_{<t}, \mathcal{G}) \log P_{\theta}(i \mid \mathbf{x}_{<t}, \mathcal{G}).
\end{equation}
A low $\mathcal{H}_l$ suggests a sharply peaked distribution, indicating that the model is confident in its prediction. Conversely, a high $\mathcal{H}_l$ signifies a flat distribution, which suggests that the model exhibits high uncertainty regarding its prediction. In practice, we consider that uncertainty exceeding the pre-defined threshold $\gamma$ at a given layer warrants intervention to recalibrate attention allocation.

\textbf{Pay Attention to Graph.} 
To mitigate GraphLLM models' tendency to overlook graph evidence, we intervene directly in the attention layers to enforce a ``\textit{Look}" operation. Unlike rigid masking or post-hoc probability scaling, our Attention Redistribution mechanism explicitly reallocates attention weights to graph tokens.

We first extract the graph token attention weights from the pre-softmax attention logits $\tilde{\mathbf{e}}_t$ for the currently generated token. When the model exhibits high uncertainty—as indicated by the layer-wise entropy $\mathcal{H}^{(l)}$ (defined in Eq. \eqref{eq:uncertainty_entropy})—we adaptively rectify the attention distribution to allocate more weight to graph tokens. The rectified attention logits are formulated as follows:

\begin{equation}
\label{eq:attention_rectification}
    \begin{gathered}
        \tilde{e}_{t,j} = e_{t,j} + \Gamma(\mathcal{H}^{(l)} > \gamma) \cdot \eta \cdot |e_{t,j}|, \\
        \textit{subject to } \quad j \in \Omega_{\mathcal{G}},
    \end{gathered}
\end{equation}

where $\eta$ is a tunable hyperparameter controlling the rectification strength, and $\Gamma(\cdot)$ denotes the gating function. The amplification is applied uniformly to all graph tokens $j\in\Omega_{\mathcal{G}}$. Since the model's final vocabulary probability distribution is derived directly from the last token's hidden state projection, we extract the last token's attention weights over the graph tokens by indexing $\tilde{e}_{t,j}$.

Following the attention intervention, we apply softmax function to redistribute attention weights across tokens when reassigning encoded hidden states. This procedure repeats autoregressively for each subsequent token prediction. Furthermore, this approach is mathematically principled as it is topology-preserving by amplifying attention proportionally to graph's intrinsic relevance.

\subsection{\underline{\textbf{\textit{Remember}}}: Graph Re-injection}
\label{sec:FFN_as_Memory_Repository}

Simply directing the model's attention to graph tokens through the ``\textit{Look}" mechanism is not enough to ensure the model's understanding of graph data. To fully enhance graph comprehension, we must extend our intervention from the attentional surface to the parametric depth, ensuring the model can ``\textit{Remember}" graph information. Specifically, we propose \textbf{Graph Re-injection}, a mechanism that dynamically injects graph information into the FFN layers based on predictive uncertainty. The overall procedure is outlined in the Appendix \ref{pseudo}.

\textbf{Memory Retrieval Formulation.} The Feed-Forward Network (FFN) layers constitute approximately two-thirds of the parameters in Transformer-based architectures, serving as the primary repository for factual knowledge~\cite{geva2020transformer}. Consider the input hidden state $x\in \mathbb{R}^d$ for a given layer, the vanilla FFN can be formulated as:
\begin{equation}
    \text{FFN}(x) = \phi(x W_1) W_2,
\end{equation}
where $\phi(\cdot)$ denotes the activation function (e.g., ReLU or SiLU), 
$W_1 \in \mathbb{R}^{d \times d_m}$ and $W_2 \in \mathbb{R}^{d_m \times d}$ represent the up-projection and down-projection weight matrices, respectively, with $d_m$ being the intermediate dimension. Specifically, the weight matrices $W_1$ and $W_2$ can be explicitly decomposed into sets of vectors as:
\begin{equation}
    \mathbf{W_1} = [\mathbf{k}_1, \dots, \mathbf{k}_{d_m}], \mathbf{W_2} = [\mathbf{v}_1, \dots, \mathbf{v}_{d_m}]^\top,
\end{equation}
where $k_i \in \mathbb{R}^d$ and $v_i \in \mathbb{R}^d$ are entries of key and value, respectively. As a consequence, FFN can be reformulated as a weighted aggregation of memory slots~\cite{pmlr-v235-jie24a}:
\begin{equation}
\label{eq: k-v ffn}
    \text{FFN}(x) = \sum_{i=1}^{d_m} \phi(\langle x, k_i \rangle) \cdot v_i,
\end{equation}
where $m_i = \phi(\langle x, k_i \rangle)$ serves as the \textit{memory coefficient}, quantifying the relevance of the $i$-th memory slot to the current input. Considering the formulation, FFN essentially performs a ``soft retrieval'' operation: the input $x$ queries the parameter space to activate relevant patterns ($k_i$) and retrieves the associated knowledge ($v_i$).

\textbf{Re-inject Graph Information.} Motivated by the findings above, we intervene in the FFN layers to implement a ``\textit{Remember}" operation following the ``\textit{Look}" stage. Specifically, we propose a \textit{Graph Re-injection} mechanism that treats graph tokens as keys and values (in Eq. \eqref{eq: k-v ffn}) and re-injects them into specific FFN layers based on uncertainty triggers.

Let $\mathbf{C}_{\mathcal{G}} = (\mathbf{g}_1, \ldots, \mathbf{g}_J) \in \mathbb{R}^{d \times J}$ denote the set of graph token embeddings, and let $x \in \mathbb{R}^d$ represent the current hidden state of the language model. We reinterpret the FFN as a key-value memory retrieval mechanism, where $x$ serves as a query vector and graph tokens act as auxiliary memory slots encoding rich graph structural context. Once the model exhibits high unertainty (i.e., $\mathcal{H}^{(l)} > \gamma$, as defined in Eq. \eqref{eq:uncertainty_entropy}), graph information is explicitly re-injected into selected FFN layers. The operation can be expressed as:
\begin{equation}
\label{eq: graph memory}
    \Delta(\mathcal{G} \mid x) = \sum_{j=1}^{J} \phi\big(\langle x, \mathbf{g}_j \rangle\big)\, \mathbf{g}_j.
\end{equation}
% \vskip -0.1in
To integrate this information, we treat the retrieved graph information as supplementary evidence that operate concurrently with the original FFN output. The calibrated output, $\widehat{\text{FFN}}(x)$, is derived by fusing the vanilla FFN output with the graph-based correction term:
\begin{equation}
\label{eq:inject graph}
    \widehat{\text{FFN}}(x) = \underbrace{(1-\alpha) \cdot \phi(x \mathbf{W}_1) \mathbf{W}_2}_{\text{Original Memory}} + \underbrace{\alpha \cdot \phi(x \mathbf{W}_1^g) \mathbf{W}_2^g}_{\text{Auxiliary Graph Memory}}.
\end{equation}
Here, Original Memory represents the vanilla FFN output, while the Auxiliary Graph Memory represents the output of our Graph Re-injection Module (defined in Eq. \eqref{eq: graph memory}). The parameters $\mathbf{W}_1^g$ and $\mathbf{W}_2^g$ are derived from graph tokens and encode supplementary graph-related information in the hidden states. The scalar $\alpha \in [0,1]$ controls the magnitude of the graph intervention.

Through \textit{Attention Redistribution} that enables the model to ``\textit{Look}" at graph information and \textit{Graph Re-injection} that enables it to ``\textit{Remember}", LoReC enhances the model's perception and understanding of the input graph structure.

\subsection{\underline{\textbf{\textit{Contrast}}}: Logit Rectification}
\label{sec:Graph_Contrastive_Decoding}

While the ``\textit{Look}" and ``\textit{Remember}" modules successfully enhance GraphLLM models' perception and understanding of graph data in latent space, the predictions may still inherit false priors. Standard decoding algorithms greedily select tokens with maximum probability, often polluted by the model's pre-training distribution rather than the specific context. In GraphLLM paradigm, this issue is further compounded by graph-specific prior biases originating from message-passing mechanisms and inductive biases inherent in graph encoders.

To eliminate these biases and enable models to yield less biased predictions, we rectify the models' vanilla logits, ensuring they can ``\textit{Contrast}" away false priors. Specifically, we introduce \textbf{Logit Rectification}, a mechanism that directly rectifies the final next-token distribution using the divergence between the vanilla output logits and two types of negative logits: text-only logit and augmented graph logit, thereby producing more faithful predictions.

\textbf{Adaptive Topological Perturbation.} To mitigate the negative prior that stems primarily from the recursive aggregation of high-degree ``hub nodes'', which dominate the message-passing trajectory and enforce a rigid homophily prior, we require the augmented graph view $\tilde{\mathcal{G}}$ that preserves dominant hub structures while stochastically pruning sparse, peripheral connections that often contain fine-grained evidence. Naive uniform edge dropout is insufficient as it fails to account for the heterogeneous importance of nodes. Following~\cite{zhu2021graph}, we employ an adaptive augmentation strategy based on degree centrality to generate $\tilde{\mathcal{G}}$. Formally, let $\varphi(v) = \log(\deg(v) + \epsilon)$ denote the degree centrality of node $v$. The connectivity strength of edge $(u,v) \in \mathcal{E}$ is defined as $s_{uv}$ and normalized to $\tilde{s}_{uv}$:
\begin{equation}
    s_{uv} = \frac{\varphi(u) + \varphi(v)}{2}, \quad \tilde{s}_{uv} = \frac{s_{\max} - s_{uv}}{s_{\max} - s_{avg}},
\end{equation}
where $s_{\max}$ and $s_{avg}$ denote the maximum and average strength values in the graph. We further introduce an overall edge drop rate $\mu$  and a truncation threshold $\tau$ (effectively preventing structural collapse). The final edge dropout distribution is formulated as:
\begin{equation}
\label{eq:adaptive_augmentation}
    w_{uv} = \min \left( \tau, \; \mu \cdot (1 - \tilde{s}_{uv}) \right).
\end{equation}
Note that the augmented graph logits ($\Psi_{\text{aug}}$) will be subtracted from the original logits (Eq. \eqref{eq:gcd_formula}). Therefore, the augmented graph view $\tilde{\mathcal{G}}$ should maximally preserve and amplify the graph inductive bias, so that the contrastive subtraction can effectively remove this bias.

\textbf{Dual-Contrastive Decoding.} To rectify text and graph biases in the original logits $\Psi_{\text{orig}}$, we obtain text-only logits $\Psi_{\text{text}}$ and perturbed-graph logits $\Psi_{\text{aug}}$ that amplify these biases, respectively. Specifically, we calculate $\Psi_{\text{text}}$ by masking all graph tokens before decoding, and calculate $\Psi_{\text{aug}}$ use the adaptively augmented graph $\tilde{\mathcal{G}}$ as input. In practice, we use a gating function $\mathbb{I}_{\text{gate}}$ that prevents invalid perturbations (when no edges are dropped or when cutting edges would remove vital information from sparse graphs). The dual-contrastive decoding process is formulated as:
    % \vskip -0.1in
\begin{equation}
\label{eq:gcd_formula}
    \Psi_{\text{final}} = \Psi_{\text{orig}} 
    + \omega \underbrace{(\Psi_{\text{orig}} - \Psi_{\text{text}})}_{\Delta_{\text{Text de-biasing}}} 
    + \beta \cdot \mathbb{I}_{\text{gate}} \underbrace{(\Psi_{\text{orig}} - \Psi_{\text{aug}})}_{\Delta_{\text{Graph de-biasing}}},
\end{equation}
where $\omega$ and $\beta$ are hyperparameters controlling the suppression magnitude. $\Delta_{\text{Text de-biasing}}$ penalizes text bias in $\Psi_{\text{text}}$ deviating from ground truth, while $\Delta_{\text{Graph de-biasing}}$ penalizes graph bias in $\Psi_{\text{aug}}$.

In practice, we define a dynamic candidate set $\mathcal{Z}_{\text{head}}$ that selectively contains only tokens with high probability under the unperturbed prediction distribution, thereby ensuring the coherence and semantic validity of the generated sequence. The complete formulation is as follows:
\begin{equation}
    \begin{aligned}
        y_t \sim \text{Softmax} & \left[ \Psi_{\text{orig}} + \omega \cdot (\Psi_{\text{orig}} - \Psi_{\text{text}}) \right. \\
        & \left. + \beta \cdot (\Psi_{\text{orig}} - \Psi_{\text{aug}}) \right], \\
        \textit{subject to } & \quad y_t \in \mathcal{Z}_{\text{head}}(y_{<t}).
    \end{aligned}
\end{equation}
Overall, the three-stage LoReC framework, comprising ``\textit{Look}", ``\textit{Remember}" and ``\textit{Contrast}", empowers GraphLLMs with enhanced graph understanding and debiased prediction capabilities, similar to human cognitive processes. The computational costs analysis is provided in Appendix \ref{time_analysis}, and the complete LoReC pseudo code is outlined in Appendix \ref{pseudo}.

\section{Experiments}

\subsection{Experimental Setup}

\begin{wraptable}{r}{0.56\textwidth} 
\vspace{-10pt} % 减少表格上方与正文的间距
  \caption{Statistics of datasets.}
  \label{tab:datasets}
  
  \begin{center}
    \begin{small}
      \begin{sc}
        % 定义列对齐：l=左对齐(文字), r=右对齐(数字), c=居中
        \begin{tabular}{lccc}
          \toprule
          Dataset  & \# Nodes & \# Edges & \# Classes \\
          \midrule
          Cora         & 2,708     & 10,556       & 7  \\
          Citeseer     & 3,327     & 9,228       & 6  \\
          PubMed    & 19,717 & 44,338 & 3 \\
          Arxiv        & 169,343   & 1,166,243   & 40 \\
          Products  & 2,449,029 & 61,859,140  & 47 \\
          \bottomrule
        \end{tabular}
      \end{sc}
    \end{small}
  \end{center}
  \vspace{-10pt} % 减少表格下方与正文的间距
\end{wraptable}

\textbf{Datasets and Evaluation Protocols.} To evaluate the performance of our proposed LoReC, we conduct experiments on five widely recognized graph datasets: PubMed~\cite{he2023harnessing}, Cora~\cite{kipf2016semi, wen2023augmenting}, Citeseer~\cite{kipf2016semi}, Ogbn-arxiv and Ogbn-products~\cite{hu2020open}. The statistics of these datasets are summarized in Table \ref{tab:datasets}. We randomly partition all datasets into training, validation, and testing sets with a ratio of 3:1:1. Note that for experiments in GraphGPT, we adhere to the its vanilla split setting with a training, validation, and testing ratio of 6:2:3. Performance is reported using Accuracy for balanced tasks and Micro-F1 for multi-label settings.

\textbf{Baseline Models.} We compare our proposed method against three distinct categories of baselines to ensure a comprehensive evaluation: (1) GNN-based models, specifically GCN ~\cite{kipf2016semi}, GAT ~\cite{velivckovic2017graph},  GKD~\cite{yang2022geometric}, and GLNN~\cite{zhang2022graphless}; (2) LLM-only approaches, including Baichuan-7b, Llama2-7b ~\cite{touvron2023llama}, Vicuna-7b~\cite{zheng2023judging} and Qwen3-8b ~\cite{yang2025qwen3} , which serve as semantic-centric baselines that rely solely on node text while ignoring graph topology; and (3) GraphLLM models, such as GraphGPT ~\cite{tang2024graphgpt} and GraphPrompter ~\cite{liu2024can} , representing state-of-the-art hybrid paradigm against which we benchmark our specific improvements.

\textbf{Implementation Details.} We utilize GraphGPT and GraphPrompter as base models. Following their original setups, we employ Vicuna-7B and Llama2-7B as backbones, respectively. Additionally, we evaluate both GraphPrompter and LoReC-GraphPrompter using a more recent and powerful backbone, Qwen3-8B. Experiments are conducted on 4 $\times$ H100 GPUs (80GB) for GraphGPT and 2 $\times$ A800 GPUs (80GB) for GraphPrompter.
%The graph encoder is a standard GNN initialized with pre-trained weights.

\textbf{Configuration.} For our ``\textit{Look}" and ``\textit{Remember}" modules, we employ the Shannon entropy of the next-token distribution to trigger attention amplification and graph re-injection, with a shared entropy threshold of $\gamma$. Specifically, we apply attention amplification across layers 15-22 with an amplification factor of $\eta$. Graph re-injection is applied to layers 8-16 with a fusion ratio of $\alpha$. For our ``\textit{Contrast}" module, we set the text-only contrast weight $\omega$ to 0.5 and the augmented graph contrast weight $\beta$ to 1.0 by default. The adaptive plausibility constraint parameter is set to 1.0 to filter out unlikely tokens. For graph view generation, we employ a degree-based edge dropout strategy with a default rate of $\mu=0.2$, which varies across datasets and selectively removes edges connected to low-degree nodes to create structurally valid contrastive views. The detailed parameter configurations can be found in the Appendix \ref{hyperparameter}.

\begin{table*}[t]
  \caption{Performance comparison based on GraphGPT. We highlight the performance of LoReC with gray background. The highest performances across different datasets are highlighted in \textbf{bold}. The results of GNN-based models and general LLMs are taken from published reports in GraphGPT.}
  \label{tab:main-results1}
  \begin{center}
  % 让表格占满页宽
  \resizebox{\textwidth}{!}{
    \begin{small} 
      \setlength{\tabcolsep}{4pt} 
      % 修改：改为9列 (lcccccccc)，去掉了第一列 Category
      \begin{tabular}{lcccccccc}
        \toprule
        % 修改：表头去掉了 Category，只保留 Model
        \multirow{2}{*}{Model} & \multicolumn{2}{c}{Arxiv-Arxiv} & \multicolumn{2}{c}{Arxiv-Pubmed} & \multicolumn{2}{c}{(Arxiv+Pubmed)-Arxiv} & \multicolumn{2}{c}{(Arxiv+Pubmed)-Cora} \\
        % 修改：cmidrule 的范围从 2-3 开始
        \cmidrule(lr){2-3} \cmidrule(lr){4-5} \cmidrule(lr){6-7} \cmidrule(lr){8-9}
         & Accuracy & Macro-F1 & Accuracy & Macro-F1 & Accuracy & Macro-F1 & Accuracy & Macro-F1 \\
        \midrule
        
        % --- GNN 部分 (作为小标题插入) ---
        \multicolumn{9}{l}{\textit{GNN-based Models}} \\
        GCN & 0.5267 & 0.3202 & 0.3940 & 0.1884 & 0.0122 & 0.0008 & 0.0187 & 0.0032 \\
        GAT & 0.5332 & 0.3118 & 0.3940 & 0.1884 & 0.1707 & 0.0285 & 0.0161 & 0.0057 \\
        GKD & 0.5570 & 0.1595 & 0.3645 & 0.2561 & 0.2089 & 0.0179 & 0.0406 & 0.0037 \\
        GLNN & 0.6088 & 0.3757 & 0.4298 & 0.3182 & 0.3373 & 0.1115 & 0.0182 & 0.0092 \\
        \midrule
        
        % --- LLM 部分 ---
        \multicolumn{9}{l}{\textit{General LLMs}} \\
        Baichuan-7b & 0.0946 & 0.0363 & 0.4642 & 0.3876 & 0.0946 & 0.0363 & 0.0405 & 0.0469 \\
        Vicuna-7b-v1.5  & 0.4962 & 0.1853 & 0.6351 & 0.5231 & 0.4962 & 0.1853 & 0.1489 & 0.1213 \\
        \midrule
        
        % --- GraphLLM 部分 ---
        \multicolumn{9}{l}{\textit{GraphLLM Models}} \\
        GraphGPT-std       & 0.6134 & 0.2607 & 0.6770 & 0.6283 & 0.6173 & 0.2594 & 0.1259 & 0.0773 \\
        GraphGPT-cot       & 0.5630 & 0.2483 & 0.5213 & 0.4816 & 0.6476 & 0.2854 & 0.1459 & 0.1287 \\
        \midrule
        
        % --- Ours 部分 (添加背景色) ---
        \multicolumn{9}{l}{\textit{Ours}} \\
        % 修改：添加 \rowcolor{gray!20}
        \rowcolor{gray!20}  LoReC(GraphGPT-std)       & \textbf{0.6312} & \textbf{0.2831} & \textbf{0.7018} & \textbf{0.6523} & 0.6428 & 0.2753 & 0.1278 & 0.0785 \\
                \rowcolor{gray!20} LoReC(GraphGPT-cot)        & 0.5826 & 0.2720 & 0.5437 & 0.5175 & \textbf{0.6734} & \textbf{0.3278} & \textbf{0.1643} & \textbf{0.1394} \\
        \bottomrule
      \end{tabular}
    \end{small}
    }
  \end{center}
  \vskip -0.1in
\end{table*}

\begin{table*}[t]
  \caption{Performance comparison based on GraphPrompter. We highlight the performance of LoReC with gray background. The highest performances across different datasets are highlighted in \textbf{bold}.}
  \label{tab:main-results2}
  \begin{center}
    \begin{small} 
    \resizebox{\textwidth}{!}{
      \setlength{\tabcolsep}{4pt} 
      % 修改：改为9列 (lcccccccc)，去掉了第一列 Category
      \begin{tabular}{lcccccccc}
        \toprule
        % 修改：表头去掉了 Category，只保留 Model
        \multirow{2}{*}{Model} & \multicolumn{2}{c}{Cora} & \multicolumn{2}{c}{Citeseer} & \multicolumn{2}{c}{Arxiv} & \multicolumn{2}{c}{Products} \\
        % 修改：cmidrule 的范围左移
        \cmidrule(lr){2-3} \cmidrule(lr){4-5} \cmidrule(lr){6-7} \cmidrule(lr){8-9}
         & Accuracy & Macro-F1 & Accuracy & Macro-F1 & Accuracy & Macro-F1 & Accuracy & Macro-F1 \\
        \midrule

        % --- LLM 部分 ---
        \multicolumn{9}{l}{\textit{General LLMs}} \\
        Llama2-7b & 0.3764 & 0.3002 & 0.3846 & 0.4731 & 0.0909 & 0.0304 & 0.1117 & 0.0474 \\
        Qwen3-8b  & 0.6568 & 0.5647 & 0.6591 & 0.6786 & 0.6204 & 0.2162 & 0.1761 & 0.1461 \\
        \midrule
        
        % --- GLM 部分 ---
        \multicolumn{9}{l}{\textit{GraphLLM Models}} \\
        GraphPrompter (Llama2-7b) & 0.7915 & 0.7134 & 0.7029 & 0.6555 & 0.7008 & 0.4153 & 0.7720 & 0.3833 \\
        GraphPrompter (Qwen3-8b)  & 0.8026 & 0.7727 & 0.6652 & 0.6430 & 0.7331 & 0.3891 & 0.7422 & 0.3675 \\
        \midrule
        
        % --- Ours 部分 ---
        \multicolumn{9}{l}{\textit{Ours}} \\
        % LoReC 行全部添加灰色背景
        \rowcolor{gray!20} LoReC (GraphPrompter-Llama2) & \textbf{0.8137} & \textbf{0.7931} & 0.7119 & 0.6418 & 0.7094 & 0.4348 & \textbf{0.7767} & \textbf{0.3854} \\
        \rowcolor{gray!20} LoReC (GraphPrompter-Qwen3)  & 0.8026 & 0.7762 & \textbf{0.7164} & \textbf{0.6804} & \textbf{0.7437} & \textbf{0.4674} & 0.7459 & 0.3772 \\
        % cora-llama-gcd:0.8081,0.7836
        % cora-qwen-gcd:0.7989,0.7705
        % citeseer-llama-gcd:0.7179,0.6490
        % citeseer-qwen-gcd:0.7134,0.6791

        \bottomrule
      \end{tabular}
      }
    \end{small}
  \end{center}
  \vskip -0.1in
\end{table*}

\subsection{Comparison with State-of-the-art Results}
We compare our proposed method against three categories of baselines: (1) GNN-based models (GCN, GAT, GKD, GLNN), (2) General LLMs (Llama2-7b, Vicuna-7b, Baichuan-7b, Qwen3-8b), and (3) GraphLLM models (GraphGPT, GraphPrompter). Note that LoReC is not trained from scratch but rather applied as a plug-and-play decoding method; thus, its performance depends on the underlying GraphLLM models. We present performance comparisons on the benchmarks utilized by GraphGPT and GraphPrompter, with results summarized in Tables~\ref{tab:main-results1} and~\ref{tab:main-results2}.

\textbf{Overall Performance.} As shown in Tables~\ref{tab:main-results1} and~\ref{tab:main-results2}, our method consistently outperforms all baselines across all datasets. Note that following GraphGPT's setup, we evaluate our method in two settings: (1) supervised learning, where models are trained and tested on the same dataset (e.g., Arxiv), and (2) zero-shot transfer, where models trained on one dataset (e.g., Arxiv) are tested on another (e.g., PubMed). In Table~\ref{tab:main-results1}, ``-v1.5-" represents version of the base Vicuna model. ``-std" and ``cot" denote the use of the standard and generated COT instruction datasets, respectively. While state-of-the-art GraphLLM models (GraphGPT and GraphPrompter) integrate graph features, their standard decoding mechanism overlooks graph information and inherits false textual priors. In contrast, our method achieves accuracy improvements of up to 2.58\% and 5.12\% over GraphGPT and GraphPrompter, respectively, with macro-F1 gains of up to 4.24\% and 7.97\%, demonstrating its effectiveness in enhancing graph understanding.

\subsection{Ablation Study}
To validate the effectiveness of three modules and the hyper-parameters we introduced in LoReC, we conducted in-depth ablation experiments as detailed below. All ablation experiments are conducted based on GraphGPT. More details can be found in the Appendix \ref{hyper}.

\begin{wrapfigure}{r}{0.5\textwidth} 
  \centering
  \vspace{-10pt} % 调整图片与上方正文的间距
  \captionsetup[subfigure]{
    labelfont=normalfont,
    size=footnotesize, % 环绕布局空间有限，建议缩小字号
    skip=2pt,
    margin=0pt
  }
  \includegraphics[width=\linewidth]{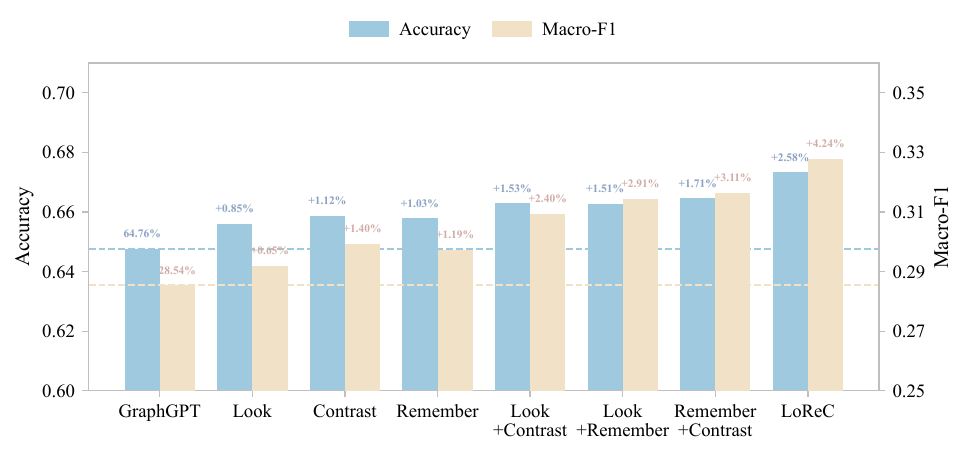}
  \caption{Impact of LoReC's components on the Arxiv dataset.}
  \label{module_ablation}
  \vspace{-10pt}
\end{wrapfigure}

\textbf{Effect of Individual Components.} To validate our three-stage design, we conduct ablation studies by enabling ``\textit{Look}", ``\textit{Remember}", and ``\textit{Contrast}" modules individually and in pairs. As illustrated in Figure \ref{module_ablation}, each individual module provides performance gains over the baseline GraphLLM model, demonstrating its independent contribution. Pairwise combinations consistently outperform individual modules, indicating synergistic effects. However, only the complete LoReC framework—integrating all three stages—achieves optimal performance, surpassing all partial configurations. These results validate our core design philosophy: ``\textit{Look}" strengthens model's perception to graph data at the attention layer, ``\textit{Remember}" extends the intervention from the attentional surface to the parametric depth by re-injecting graph evidence into FFN layers, and ``\textit{Contrast}" finally rectifies false priors through dual-contrastive decoding. The complementary nature of these three stages is essential for comprehensively enhancing GraphLLM models' understanding of graph data.

\begin{figure}[ht]
  \centering
  % 针对单栏微调字号和间距
  \captionsetup[subfigure]{labelfont=normalfont, size=small, skip=2pt}
  
  \begin{subfigure}[b]{0.35\linewidth} % 缩小比例，节省垂直空间
    \centering
    \includegraphics[width=\linewidth]{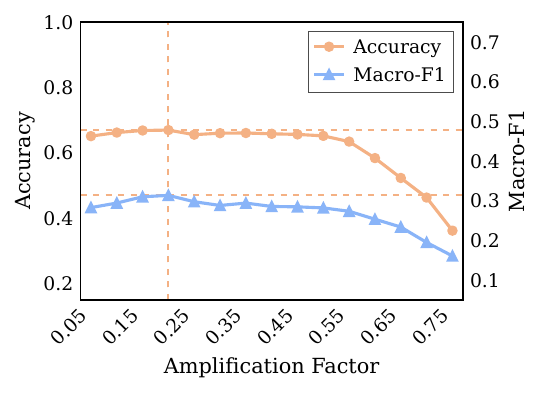}
    % \caption{a} % 生成 (a) 标识
    \label{fig:ablation_attn}
  \end{subfigure}
  \hfil % 使用弹性间距
  \begin{subfigure}[b]{0.35\linewidth}
    \centering
    \includegraphics[width=\linewidth]{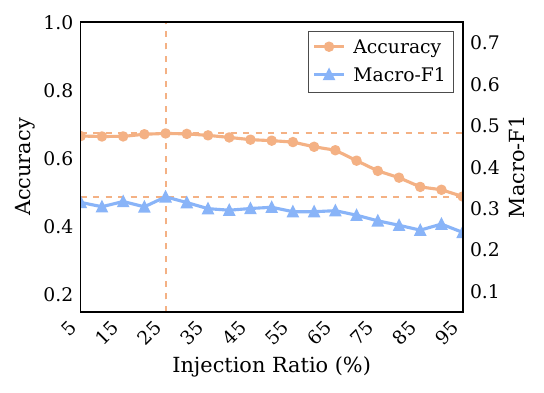}
    % \caption{b} % 生成 (b) 标识
    \label{fig:ablation_ffn}
  \end{subfigure}
  \caption{Ablation studies on the Arxiv dataset: (Left) Results under different amplification factors $\eta$; (Right) Results under different injection ratios $\alpha$. Both use GraphGPT as the base model.}
  \label{ablation}
  % \vspace{-10pt} % 仅在空间极度受限时使用，NeurIPS 默认间距通常很优雅
\end{figure}

\textbf{Effect of Amplification Factor and Injection Ratio.} As shown in Figure \ref{ablation}, we analyze the impact of  amplification factors $\eta$ and injection ratios $\alpha$ on both accuracy and macro-F1 scores. For $\eta$, values between 0.05 and 0.45 consistently improve model performance, with the optimal setting observed at approximately 0.20. Similarly, injection ratio $\alpha$ in the moderate range of $15\%$ to $35\%$ yields positive effects, peaking at around $25\%$. Notably, performance degrades when $\alpha$ exceeds $35\%$, indicating that excessive graph information injection can be detrimental.

\section{Conclusion}
In this paper, we explain why LLMs struggle in GraphLLM paradigm and propose LoReC, a novel decoding method that comprehensively enhances LLMs' understanding of graph data through three stages: ``\textit{Look}", ``\textit{Remember}", and ``\textit{Contrast}". LoReC operates as a plug-and-play method requiring no additional fine-tuning, enabling seamless integration with existing GraphLLM models. Extensive experiments across multiple datasets demonstrate its effectiveness. 

% \newpage
{
\small
\bibliographystyle{IEEEtran} % 指定引用风格为 IEEEtran
\bibliography{main} % 这里的文件名不要带 .bib 后缀
}
\newpage
% \input{checklist.tex}

% \section*{Impact Statement}
% This work aims to improve LLMs' ability to understand graph-structured data, which has broad applications in social networks and molecular analysis. Our training-free approach reduces computational costs and makes advanced graph learning more trustworthy. While applicable to sensitive areas like social networks, our work focuses on methodological improvements rather than specific applications.


%%%%%%%%%%%%%%%%%%%%%%%%%%%%%%%%%%%%%%%%%%%%%%%%%%%%%%%%%%%%%%%%%%%%%%%%%%%%%%%
%%%%%%%%%%%%%%%%%%%%%%%%%%%%%%%%%%%%%%%%%%%%%%%%%%%%%%%%%%%%%%%%%%%%%%%%%%%%%%%
% APPENDIX
%%%%%%%%%%%%%%%%%%%%%%%%%%%%%%%%%%%%%%%%%%%%%%%%%%%%%%%%%%%%%%%%%%%%%%%%%%%%%%%
%%%%%%%%%%%%%%%%%%%%%%%%%%%%%%%%%%%%%%%%%%%%%%%%%%%%%%%%%%%%%%%%%%%%%%%%%%%%%%%
\newpage
\appendix

\section*{Technical appendices and supplementary material}
In the Appendix, we provide supplementary material to the main paper. The structure is as follows:
\begin{enumerate}
    \item Section A presents the attention distribution across graph and text tokens on Pubmed dataset.
    \item Section B presents computational costs analysis of LoReC.
    \item Section C outlines the experimental settings and hyper-parameters analysis in detail.
    \item Section D presents pseudo codes of LoReC.
\end{enumerate}

\section{Attention Distribution Across Graph and Text Tokens on PubMed Dataset.}
\label{pubmed}
\vskip 0.1in
\begin{figure}[ht]
  \centering
  \captionsetup[subfigure]{labelfont=normalfont}
  \begin{subfigure}[b]{0.45\linewidth}
    \centering
    \includegraphics[width=\linewidth]{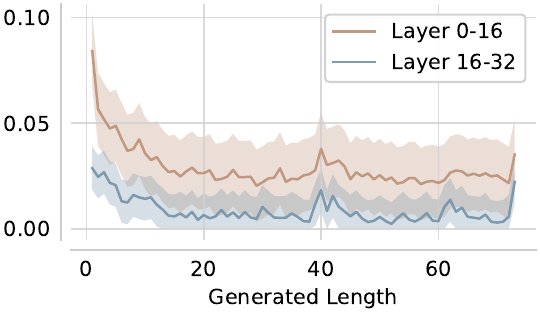}
    \caption{Graph Tokens}
  \end{subfigure}%
  \hspace{5pt}
  \begin{subfigure}[b]{0.45\linewidth}
    \centering
    \includegraphics[width=\linewidth]{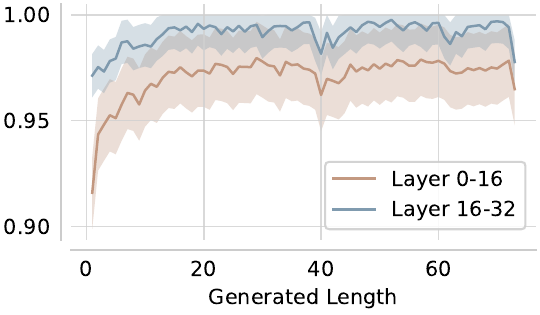}
    \caption{Text Tokens}
  \end{subfigure}
\vskip 0.1in
  \caption{Visualization of the attention distribution between graph and text tokens across decoding layers on the PubMed dataset. The results illustrate the dynamic evolution of attention patterns as the generation proceeds.}
  \label{attn_figure}
\end{figure}

As illustrated in \cref{attn_figure}, our experiments using GraphGPT on the PubMed dataset reveal that attention progressively shifts from graph tokens to text tokens as generation proceeds. This phenomenon explains the suboptimal performance of GraphLLM models: due to their autoregressive nature, LLMs tend to over-rely on textual features while neglecting graph information. Notably, this attention imbalance becomes more pronounced in deeper layers, confirming our hypothesis.

\section{Computational Costs Analysis}
\label{time_analysis}
LoReC operates ``Look" and ``Remember" dynamically based on uncertainty. Specifically, when uncertainty remains low across all layers, indicating high model confidence, these operations are not triggered. This mechanism efficiently enhances model's comprehension of graph data without extra computation. Similarly, for the ``Contrast" operation, we employ a gate function $\mathbb{I}_\text{gate}$ to dynamically trigger graph-contrast logits calculation only when needed, thereby optimizing both computational time and memory consumption. Overall, while LoReC incurs a modest increase in time and memory costs, it achieves substantial performance improvements. The comparisons on latency, time cost, and memory are shown in Table~\ref{tab:time}.

% \vskip 0.1in
\begin{table}[ht]
  \centering
  \caption{Performance comparison of different methods and LoReC in latency, time cost, memory usage, and accuracy.}
  \label{tab:time}
  \vspace{8pt} 
  
  % 1. 使用 small 或 footnotesize 确保字号专业
  \small 
  % 2. 手动调整列间距，避免表格太窄或太宽 (默认是 6pt，可调为 10pt 增加呼吸感)
  \setlength{\tabcolsep}{12pt} 
  
  \begin{tabular}{lcccc}
    \toprule
    {Method} & \makecell{{Latency} \\ (ms/token)} & \makecell{{Memory} \\ (GB)} & \makecell{{Time Cost(s)}} & \makecell{{Accuracy(\%)}} \\
    \midrule
    LoReC         & 82.69 & 27.93 & 280.65 & 71.64 \\
    Qwen3-8b      & 13.67 & 18.22 & 259.15 & 65.91 \\
    GraphPrompter & 45.23 & 18.43 & 201.11 & 66.52 \\
    \bottomrule
  \end{tabular}
\end{table}

\newpage

\section{Hyper-parameters Analysis}
\label{hyper}
\subsection{Complete Configurations}
\label{hyperparameter}
Following GraphGPT~\cite{tang2024graphgpt} and GraphPrompter~\cite{liu2024can}, we adopt their reported parameter initializations and only optimize the hyperparameters specific to LoReC. For GraphGPT, the edge dropout probability $\mu$ ranges from 0.2 to 0.3 to maintain essential graph structural integrity. Additionally, we adaptively change the edge threshold across different datasets to avoid removing vital information from sparse graphs. All detailed dataset-specific hyperparameter configurations are summarized in Table~\ref{tab:parameters}.

% \vskip 0.1in
\begin{table}[ht]
  \caption{Parameter settings for LoReC across different base models and datasets.}
  \label{tab:parameters}
  \vspace{8pt}
  \centering
  \small % 减小基础字号以适应多列布局
  \resizebox{\textwidth}{!}{
    \begin{tabular}{llcccccccc}
      \toprule
      {Base Model} & {Dataset} & $\mu$ & $\tau$ & $\omega$ & $\beta$ & $\eta$ & $\alpha$ & \makecell{{Entropy} \\ {Threshold}} & \makecell{{Edge} \\ {Threshold}} \\
      \midrule
      \multirow{3}{*}{GraphGPT}  
       & Arxiv    & 0.2 & 0.7 & 0.5 & 1.0 & 0.2  & 0.25 & 0.75 & 10 \\
       & PubMed   & 0.2 & 0.7 & 1.0 & 1.0 & 0.1  & 0.1  & 0.75 & 10 \\
       & Cora     & 0.3 & 0.7 & 0.8 & 1.0 & 0.15 & 0.2  & 0.75 & 20 \\
      \midrule
      \multirow{4}{*}{GraphPrompter} 
       & Cora     & 0.2 & 0.7 & 1.0 & 1.0 & 0.1  & 0.1  & 0.75 & 10 \\
       & Citeseer & 0.2 & 0.7 & 1.0 & 1.0 & 0.1  & 0.1  & 0.75 & 10 \\
       & Arxiv    & 0.2 & 0.7 & 1.0 & 1.0 & 0.1  & 0.1  & 0.75 & 10 \\
       & Products & 0.2 & 0.7 & 1.0 & 1.0 & 0.1  & 0.1  & 0.75 & 10 \\
      \bottomrule
    \end{tabular}
  }
\end{table}
% \vskip 0.1in

\subsection{Effect of Dual-contrastive Decoding Strength.}
% \label{appendix_figure2}

\begin{figure}[ht]
  \centering
  % 针对四图并排微调样式
  \captionsetup[subfigure]{
    labelfont=normalfont,
    size=footnotesize, % 缩小标签字号
    skip=2pt
  }
  % 1. appendix_cd.pdf
  \begin{subfigure}[b]{0.23\linewidth}
    \centering
    \includegraphics[width=\linewidth]{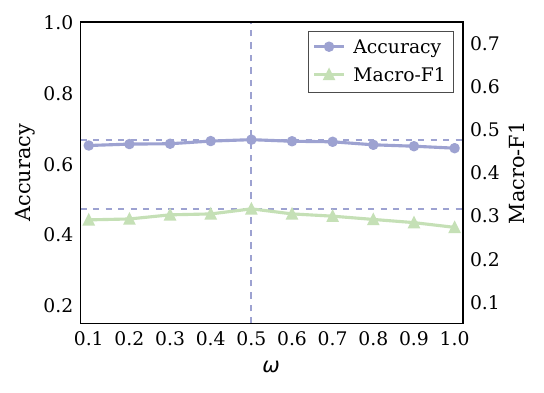}
    \caption{} % 生成 (a)
    \label{fig:cd}
  \end{subfigure}
  \hfill
  % 2. appendix_cd2.pdf
  \begin{subfigure}[b]{0.23\linewidth}
    \centering
    \includegraphics[width=\linewidth]{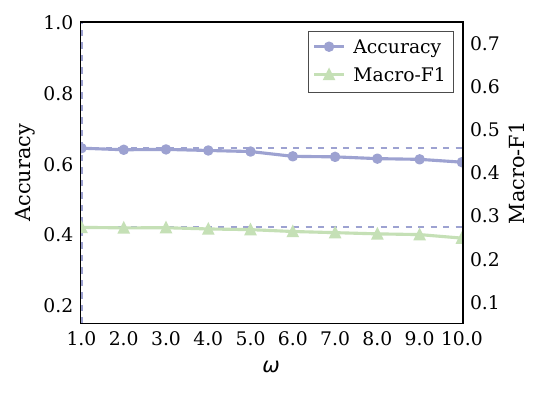}
    \caption{} % 生成 (b)
    \label{fig:cd2}
  \end{subfigure}
  \hfill
  % 3. appendix_cg.pdf
  \begin{subfigure}[b]{0.23\linewidth}
    \centering
    \includegraphics[width=\linewidth]{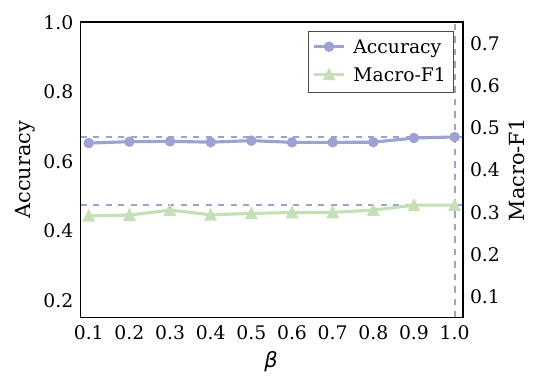}
    \caption{} % 生成 (c)
    \label{fig:cg}
  \end{subfigure}
  \hfill
  % 4. appendix_cg2.pdf
  \begin{subfigure}[b]{0.23\linewidth}
    \centering
    \includegraphics[width=\linewidth]{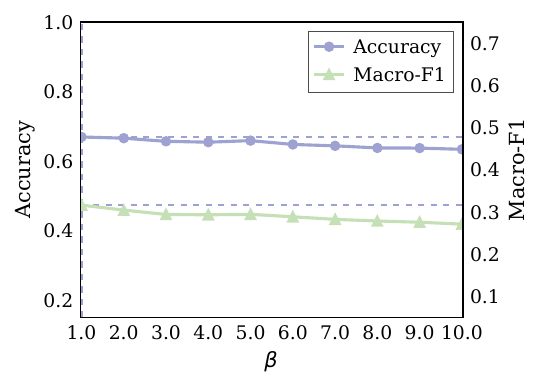}
    \caption{} % 生成 (d)
    \label{fig:cg2}
  \end{subfigure}

  \caption{(a-b) Results under different text-contrastive magnitudes $\omega$; (c-d) Results under different graph-contrastive magnitudes $\beta$.}
  \label{appendix_gcd_combined}
\end{figure}

 To identify optimal settings for the contrastive weights $\omega$ and $\beta$, we perform a grid search over the range [0, 10.0]. As illustrated in \cref{fig:cd2} and \cref{fig:cg2}, we observe that while moderate contrastive strength enhances performance, excessive values ($> 1.0$) can compromise the model's original inference capability by over-correcting the logits. Specifically, the contrastive divergence terms tend to overwhelm the final logit distribution, causing severe over-correction that distorts the model's original inference capability. Therefore, we conduct a refined grid search over the range [0.1, 1.0] to determine optimal values. As demonstrated in \cref{fig:cd} and \cref{fig:cg}, the optimal text-contrastive magnitude is approximately 0.5, while the optimal graph-contrastive magnitude is around 1.0.

\subsection{Effect of Entropy Threshold and Edge Threshold.}
\begin{figure}[ht]
  \centering
  \captionsetup[subfigure]{labelfont=normalfont}
  \begin{subfigure}[b]{0.45\linewidth}
    \centering
    \includegraphics[width=\linewidth]{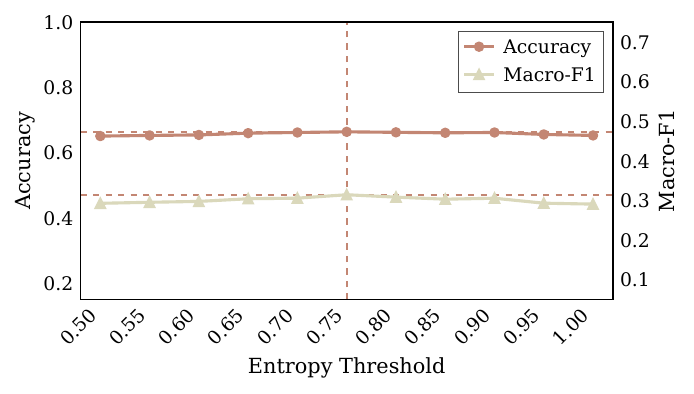}
  \end{subfigure}%
  \hspace{5pt}
  \begin{subfigure}[b]{0.45\linewidth}
    \centering
    \includegraphics[width=\linewidth]{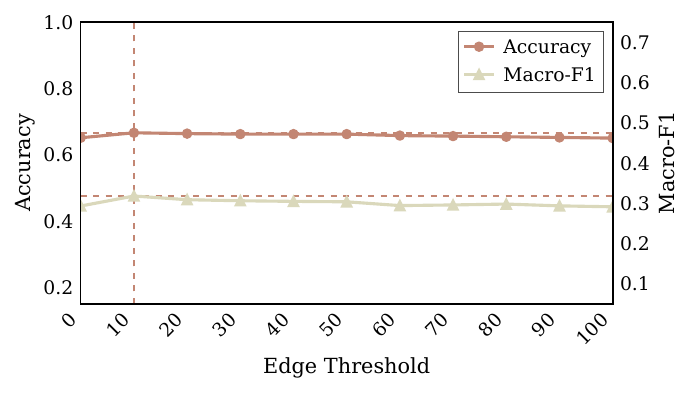}
  \end{subfigure}

  \caption{(Left) Results under different entropy threshold. (Right) Results under different edge threshold. Both evaluated on the Arxiv dataset with GraphGPT as the base model.}
  \label{threshold}
    % \vskip -0.2in
\end{figure}

Figure~\ref{threshold} demonstrates the impact of entropy threshold and edge threshold. For the entropy threshold, which determines when to activate the ``\textit{Look}" and ``\textit{Remember}" modules, exhibits optimal performance around 0.75 within an effective range of [0.65, 0.90]. The edge threshold serves as the gating function $\mathbb{I}_{\text{gate}}$ in Eq. \eqref{eq:gcd_formula} to determine whether graph-contrastive decoding should be applied, which achieves peak performance at approximately 10 edges. Understandably, a high edge threshold makes graph-contrastive decoding hard to trigger, while a low edge threshold would remove vital information from sparse graphs.

\newpage

\section{Pseudo Codes of LoReC}
We present the algorithm pipeline as follows. The equations mentioned in the algorithm can be seen in the main text.
\label{pseudo}

\vskip 0.1in
\begin{algorithm}[htb]
\caption{\textit{Look}: Attention Redistribution}
\label{alg:attention_redistribution}
\begin{algorithmic}[1]
\STATE {\bfseries Input:} Graph tokens $\mathbf{C}_{\mathcal{G}}$, Text tokens $\mathbf{C}_{\mathcal{T}}$, Model $\mathcal{M}$.
\STATE At every decoding step $t$:
\STATE Initial set \textit{trigger} = False.
\FOR{$l = 0$ {\bfseries to} $L-1$}
    \STATE $\mathcal{H}_t^{(l)} = - \frac{1}{\log N} \sum_{i=1}^{N} p_{\theta}^{(l)} 
    \log p_{\theta}^{(l)}$ (Eq. \eqref{eq:uncertainty_entropy}).
    \IF{$\mathcal{H}_t^{(l)} > \gamma$}
        \STATE \textit{trigger} = True.
    \ENDIF
\ENDFOR
\FOR{$l = l+1$ {\bfseries to} $L-1$}
    \STATE Select graph attention logits $\mathbf{e}_j^{(l)}$ at layer $l$.
    \STATE Execute $\tilde{e}_{j}^{(l)} = e_{j}^{(l)} + \eta\,|e_{j}^{(l)}|$ (Eq. \eqref{eq:attention_rectification}).
    \STATE Recalculate attention weights $\tilde{\mathcal{A}}^{(l)} =\text{Softmax}(\tilde{\mathbf{e}}^{(l)})$.
\ENDFOR
\STATE {\bfseries Output:} Redistributed attention weights $\tilde{\mathcal{A}}$.
\end{algorithmic}
\end{algorithm}

\vskip 0.1in

\begin{algorithm}[htb]
\caption{\textit{Remember}: Graph Re-injection}
\label{alg:graphvr_redistribution}
\begin{algorithmic}[1]
\STATE {\bfseries Input:} 
    Input embedding $x^{(0)}$, 
    Graph Tokens $\mathbf{C}_{\mathcal{G}}$.
\STATE Initial set \textit{trigger} = False.
\FOR{$l = 0$ {\bfseries to} $L-1$}
    \STATE $\mathbf{FFN}_{\text{base}} = \phi(\mathbf{x}^{(l)} \mathbf{W}_1^{(l)}) \mathbf{W}_2^{(l)}$;

    \STATE $\mathcal{H}_t^{(l)} = - \frac{1}{\log N} \sum_{i=1}^{N} p_{\theta, i}^{(l)} \log p_{\theta, i}^{(l)}$  \quad (Eq. (\ref{eq:uncertainty_entropy})).
    
    \IF{$\mathcal{H}_t^{(l)} > \gamma$ and \textit{trigger} == False}
        \STATE \textit{trigger} = True.
        \STATE $\mathbf{W}_1^g \leftarrow \mathbf{C}_{\mathcal{G}}, \quad \mathbf{W}_2^g \leftarrow \mathbf{C}_{\mathcal{G}}^\top$;
        \STATE $\mathbf{Mem}_{\text{graph}} = \phi(\mathbf{x}^{(l)} \mathbf{W}_1^g) \mathbf{W}_2^g$;
    
        \STATE $\mathbf{FFN}_{\text{final}} = (1-\alpha) \cdot \mathbf{FFN}_{\text{base}} + \alpha \cdot \mathbf{Mem}_{\text{graph}}$.
    \ELSE
        \STATE $\mathbf{FFN}_{\text{final}} = \mathbf{FFN}_{\text{base}}$.
    \ENDIF
\ENDFOR

\STATE {\bfseries Output:} Final hidden state $x^{(L)}$.
\end{algorithmic}
\end{algorithm}

\begin{algorithm}[htb]
   \caption{LoReC Strategy}
   \label{alg:lorec_inference}
\begin{algorithmic}[1]
   \STATE {\bfseries Input:} Graph tokens $\mathbf{C}_{\mathcal{G}}$, Text tokens $\mathbf{C}_{\mathcal{T}}$, Model $\mathcal{M}$.
   \STATE Construct perturbed graph $\tilde{\mathcal{G}}$ via Adaptive Augmentation (Eq. \eqref{eq:adaptive_augmentation}).
   \STATE At every decoding step $t$:
   \FOR{$l = 0$ {\bfseries to} $L-1$}
       \STATE Compute Uncertainty $\mathcal{H}_t^{(l)}$ (Eq. \eqref{eq:uncertainty_entropy}).
       \IF{$\mathcal{H}_t^{(l)} > \gamma$}
           \STATE Apply \textbf{Look}: Reallocate Attention (Eq. \eqref{eq:attention_rectification}).
           \STATE Apply \textbf{Remember}: Inject Graph (Eq. \eqref{eq:inject graph}).
       \ENDIF
       \STATE Calculate logits $\mathbf{v}_{\text{orig}} = \mathcal{M}(\mathcal{G}, \mathbf{x}_{<t})$.
       \STATE Calculate logits $\mathbf{v}_{\text{text}} = \mathcal{M}(\emptyset, \mathbf{x}_{<t})$.
       \STATE Calculate logits $\mathbf{v}_{\text{aug}} = \mathcal{M}(\tilde{\mathcal{G}}, \mathbf{x}_{<t})$.
       \STATE Apply \textbf{Contrast}: $\mathbf{v}_{\text{final}} \leftarrow \mathbf{v}_{\text{orig}}, \mathbf{v}_{\text{text}}, \mathbf{v}_{\text{aug}}$ (Eq. \eqref{eq:gcd_formula}).
       \STATE Sample $y_t \sim \text{Softmax}(\mathbf{v}_{\text{final}})$.
       \STATE $\mathbf{x}_{<t+1} \leftarrow \text{Concat}(\mathbf{x}_{<t}, y_t)$.
   \ENDFOR
   \STATE {\bfseries Output:} Model prediction $y_t$
\end{algorithmic}
\end{algorithm}

%%%%%%%%%%%%%%%%%%%%%%%%%%%%%%%%%%%%%%%%%%%%%%%%%%%%%%%%%%%%

%%%%%%%%%%%%%%%%%%%%%%%%%%%%%%%%%%%%%%%%%%%%%%%%%%%%%%%%%%%%

\end{document}